\documentclass[letterpaper]{article} 
\usepackage{aaai2026}  
\usepackage{times}  
\usepackage{helvet}  
\usepackage{courier}  
\usepackage[hyphens]{url}  
\usepackage{graphicx} 
\usepackage{natbib}  
\usepackage{caption} 
\usepackage{algorithm}
\usepackage{algorithmic}
\usepackage{booktabs}
\usepackage{bm}
\usepackage{amsmath}
\usepackage{multirow}

\newtheorem{definition}{Definition}

\newcommand{\dbox}[1]{\makebox[1.5em][r]{#1}}

\usepackage{newfloat}
\usepackage{listings}
\DeclareCaptionStyle{ruled}{labelfont=normalfont,labelsep=colon,strut=off} 
\floatstyle{ruled}
\newfloat{listing}{tb}{lst}{}
\floatname{listing}{Listing}
\title{Efficient Thought Space Exploration Through Strategic Intervention}
\author{
    Ziheng Li\textsuperscript{\rm 1}\thanks{Work done during internship at Baidu.}
    Hengyi Cai\textsuperscript{\rm 2},
    Xiaochi Wei\textsuperscript{\rm 2},
    Yuchen Li\textsuperscript{\rm 2},\\
    Shuaiqiang Wang\textsuperscript{\rm 2},
    Zhi-Hong Deng\textsuperscript{\rm 1}\thanks{Corresponding Author.},
    Dawei Yin\textsuperscript{\rm 2}
}
\affiliations{
    \textsuperscript{\rm 1}State Key Laboratory of General Artificial Intelligence, School of Intelligence Science and Technology, Peking University\\
    \textsuperscript{\rm 2}Baidu Inc.\\
    \{liziheng, zhdeng\}@pku.edu.cn, caihengyi@ict.ac.cn, weixiaochi@baidu.com, \\
    \{shqiang.wang, yuchenli1230\}@gmail.com, yindawei@acm.org\\
}

\nocopyright

\begin{document}

\maketitle

\begin{abstract}
While large language models (LLMs) demonstrate emerging reasoning capabilities, current inference-time expansion methods incur prohibitive computational costs by exhaustive sampling.
Through analyzing decoding trajectories, we observe that most next-token predictions align well with the golden output, except for a few critical tokens that lead to deviations.
Inspired by this phenomenon, we propose a novel Hint-Practice Reasoning (HPR) framework that operationalizes this insight through two synergistic components: 1) a \textit{hinter} (powerful LLM) that provides probabilistic guidance at critical decision points, and 2) a \textit{practitioner} (efficient smaller model) that executes major reasoning steps. 
The framework's core innovation lies in \textbf{Distributional Inconsistency Reduction (DIR)}, a theoretically-grounded metric that dynamically identifies intervention points by quantifying the divergence between practitioner's reasoning trajectory and hinter's expected distribution in a tree-structured probabilistic space. 
Through iterative tree updates guided by DIR, HPR reweights promising reasoning paths while deprioritizing low-probability branches. 
Experiments across arithmetic and commonsense reasoning benchmarks demonstrate HPR's state-of-the-art efficiency-accuracy tradeoffs: it achieves comparable performance to self-consistency and MCTS baselines while decoding only 1/5 tokens, and outperforms existing methods by at most 5.1\% absolute accuracy while maintaining similar or lower FLOPs.
\end{abstract}


\section{Introduction}
Large language models (LLMs) have made significant advancements in solving complex reasoning tasks~\citep{openai_gpt-4_2023, wei_chain--thought_2022, kojima_large_2022, huang_towards_2023}, such as mathematical problems~\citep{cobbe_training_2021} and code generation~\cite{chen_program_2023}. 
The emergence of these reasoning abilities comes from the scaled pre-training \citep{kaplan_scaling_2020, openai2024openaio1card,deepseekai2025deepseekr1incentivizingreasoningcapability} and curated prompting strategies for eliciting reasoning~\citep{wei_chain--thought_2022, kojima_large_2022, chen_program_2023, wang_strategic_2024, li2025aisearchparadigm,zheng_progressive-hint_2024}. 

Recently, researchers have found that leveraging curated decoding strategies to scale LLM inference-time reasoning expansion can also lead to notable performance gains~\citep{zhang_what_2025, wang_chain--thought_2024}.
Typical strategies include extensive sampling, structured rationale topology, and external guidance.
For example, self-consistency~\citep{wang_self-consistency_2022} and CoT-decoding~\citep{wang_chain--thought_2024} sample multiple rationales and select the most consistent answer, Best-of-N~\citep{gui_bonbon_2024} uses a score model to select the most reasonable thought from all generations, Tree-of-thoughts (ToT)~\citep{yao_tree_2023} and RAP~\citep{hao_reasoning_2023} formalize thought exploration as a tree-structure search problem which allows looking ahead and backtracking to make optimal choices, and AdaSwitch~\citep{sun_adaswitch_2024} introduces the collaboration with a stronger LLM to overcome the hard part of the reasoning.
However, current reasoning exploration methods leave three unsolved issues: (1) Low token efficiency: pure-sampling methods cannot reuse correct prefixes while tree-structured methods generate numerous intermediate branches but only a few of them can lead to the final answer. Furthermore, when dealing with complex problems, these methods require sampling prohibitively many trajectories to find a correct path. (2) Local exploration guidance: heuristic-based search algorithm easily converges to a ``good'' path and fails to explore the whole thought space. Although MCTS uses UCT to balance exploration and exploitation, it does not fully exploit token probabilities, resulting in an underestimation of truly valuable branching points. (3) High external dependence: to verify generated steps or overcome overly hard reasoning steps, a common practice is to apply an ad-hoc trained verifier or considerable usage of a stronger LLM reasoner, which will introduce considerable extra cost.

\begin{figure}[t]
    \centering
    \includegraphics[width=0.97\linewidth]{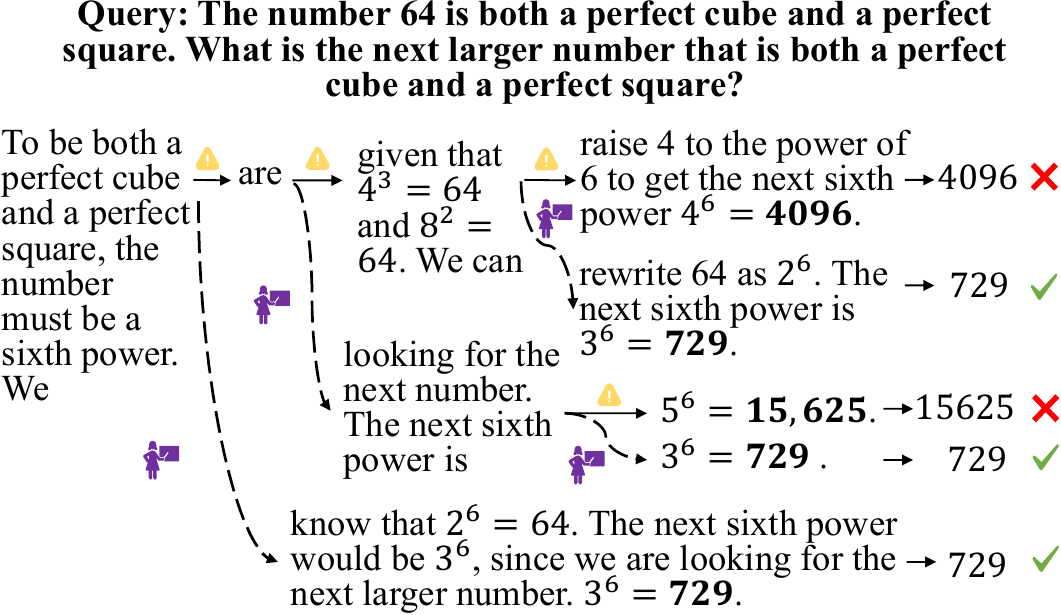}
    \caption{Illustration of the HPR framework. In each iteration, hinter will analyze existing trajectories and find the most promising intermediate state to create an alternative branch. Practitioner then complete the new branch.}
    \label{fig:example}
\end{figure}

To address these issues, we draw inspiration from human problem solving where students often benefit significantly from just a handful of advisor's hints, especially for complex problems.
Such timely ``nudges'' can reveal underexplored directions and steer a learner onto the correct trajectory without requiring extensive trial and error.
Inspired by this phenomenon, we investigate two questions in this paper: (1) whether LLMs can similarly benefit from such a few well-placed short ``hints'' to conduct efficient reasoning path exploration. (2) How to place hints into the regular generation to provide global and efficient exploration guidance.

For the first question, we conduct an analytical experiment by providing the model with a correct chain-of-thought as the reference and then identifying where its predictions deviate.
Our findings show that within the chain-of-thought decoding, an 3B model's next-token predictions align with those of a 32B model for the majority of tokens, only a small fraction cause significant deviations in reasoning. 
These sparse ``critical tokens'' therefore represent prime opportunities for targeted intervention: although it involves a larger model at strategic points, the low intervention frequency yields lower costs than either exhaustive sampling with the smaller model or directly using the larger model.

Building upon the empirical observation that strategic intervention at sparse critical tokens can effectively redirect reasoning trajectories at minimal cost, we accordingly propose a novel Hint-Practice Reasoning (HPR) framework (illustrated in Figure~\ref{fig:example}), which implements this insight via two distinct roles: (1) a hinter (powerful but costly model) provides targeted hints at sparse critical tokens; (2) a practitioner (efficient smaller model) executes the majority of reasoning steps. Different from existing multi-agent studies~\citep{sun_adaswitch_2024,gui_bonbon_2024}, HPR hinter selectively corrects the critical mistakes of the practitioner and additionally guides exploration, which reduces the large model's generation burden.
The left challenge lies in identifying where to apply interventions.
Intuitively, the practitioner should think twice when it significantly deviates from hinter's prediction. Furthermore, we expect the practitioner to start exploring from a promising and uncertain (from the global view) partial trajectory to reuse computation and maximize exploration.

We ground this heuristic into a unified probabilistic theory, which characterizes practitioner's reasoning tree by a distribution and induces \textbf{Distributional Inconsistency Reduction (DIR)}, a novel metric that quantifies the change of discrepancy between the whole reasoning tree (rather than a path) and hinter's expected distribution.
Generally, we formalize the reasoning process as a tree-structured probabilistic space where each node represents a partial reasoning trajectory. The final answer is derived through aggregation operations (e.g., weighted voting) on leaf nodes. 
The hinter's interventions essentially induce tree updates to this structure: when providing hints at critical tokens, the practitioner's subsequent predictions are dynamically reweighted based on the hinter's distributional guidance, effectively deprioritizing low-probability branches while amplifying promising trajectories.
Concretely, during each iteration to update the tree-structured reasoning trajectory, HPR first analyzes the generated trajectories, employing an adaptive intervention strategy that prioritizes nodes with maximal DIR values (indicating where the practitioner's predictions diverge most severely from the hinter's guidance), and then activates the hinter to construct a piece of hint.
The practitioner subsequently propagates these hints to complete the remaining generation.
This iterative refinement terminates when the maximum iteration count is reached.

Our contributions are threefold:
\begin{itemize}
    \item 
    We propose HPR, a cross-reasoner collaboration decoding framework for LLM reasoning. Our method provides a new way to organize the reasoning thought space exploration.
    \item 
    We establish a theoretical guidance \textbf{DIR} that unifies path exploration quality and hinter-practitioner distribution alignment, providing principled guidance for reasoning tree updating.
    \item 
    Comprehensive experiments across arithmetic and commonsense reasoning tasks demonstrate HPR's state-of-the-art efficiency-accuracy tradeoff, requiring only 1/5 tokens to get comparable performance against baselines, and improving the performance by at most 5.1\% while maintaining a similar or lower FLOPs.
\end{itemize}

\section{Related Work}
\subsection{Inference-time LLM Reasoning}
Recent studies have found increasing inference-time computations can improve the output quality of LLMs, particularly for complex tasks that needs long-chain reasoning~\citep{wu_inference_2024, brown_large_2024, fu_efficiently_2024, chu_navigate_2024}.
Chain-of-thought prompting~\citep{wei_chain--thought_2022} is the first to demonstrate intermediate rationales can boost the reasoning ability of LLMs.
Zero-shot CoT~\citep{kojima_large_2022} employs a reasoning instruction to activate the rationale generation and avoid the dependence on few-shot demonstrations.
Subsequent research focuses on further improving the quality of the rationale.
UAG~\citep{yin_reasoning_2024} identifies the risky steps of the reasoning path by the uncertainty and adjust demonstrations to eliminate the risk.
\citet{zhou_least--most_2023, press_measuring_2023} prompt the LLM to decompose the query into easier sub-parts.
\citet{madaan_self-refine_2023,manakul_selfcheckgpt_2023} self-correct the reasoning mistakes by tasks-specific prompting. \citet{kumar2024traininglanguagemodelsselfcorrect} internalized the self-correction ability by reinforcement learning.
\citet{cobbe_training_2021, wang_math-shepherd_2024, li_making_2023} propose to train a verifier to select the optimal rationale. Our approach exploit the potentials of autoregressive decoding, which is task-agnostic and does not require curated demonstration, prompts, nor ad-hoc trained models.
\subsection{Thought Space Exploration}
The autoregressive decoding nature of LLM poses challenges in finding the globally optimal reasoning path. To address this, researchers utilized the probabilistic modeling nature of LLMs to explore diverse thoughts. Self-consistency~\citep{wang_self-consistency_2022} adopts a sample-and-marginalize decoding strategy, finding the most consistent answer among multiple sampled paths. CoT-decoding~\citep{wang_chain--thought_2024} create diverse paths by enumerating the top-K candidates of the starting token and select the most plausible path by comparing answer's confidence score. Further advancing this line of research, Tree-of-thoughts and Graph-of-Thoughts~\citep{yao_tree_2023, long_large_2023, besta_graph_2024, mo_tree_2024, zhang_thought_2024} generalizes the topology of rationale into the tree or graph structures by allowing LLM to try multiple reasoning branches at each step. RAP~\citep{hao_reasoning_2023} regard reasoning as planning and introduce Monte Carlo Tree Search (MCTS) to organize the reasoning exploration space, which iteratively expand new paths under the guide of reward models.
AdaSwitch~\citep{sun_adaswitch_2024} formulates rationale generation as a multi-agent task. When the local model finds an error in the reasoning process, it asks for a more powerful cloud model to correct the mistake.
Our approach follows tree-structure thought exploration paradigm. We achieve more efficient reasoning by delegating part of generation to a stronger reasoner and further exploiting LLM's probabilistic information to organize the tree growth.

\section{Sparse Disparity in LLM Reasoning}
We start by briefly introducing the reasoning procedure for an LLM from a probabilistic view. Given a query $\bm x$, our goal is to predict the answer's distribution $P_\phi(\bm y|\bm x)$ using an autoregressive model parameterized by $\phi$. For complex problems, answer $y$ cannot be directly inferred from the query. We need to first infill the intermediate rationale $\bm r=r_1r_2\cdots r_l$ and then get the final answer:
\begin{equation}
    P_\phi(\bm y|\bm x)=\sum_{\bm r}P_\phi(\bm r|T(\bm x))P_\phi(\bm y|T(\bm x), \bm r),
\end{equation}
where $T(\cdot)$ represents a prompt template (e.g. appending ``\textit{Let's think step by step}'' to the end of $\bm x$). For simplicity, we omit $T(\cdot)$ in the remaining paper.
Although generating rationales is generally easier than directly generating the answer, LLMs can still deviate from the correct reasoning path, particularly for smaller models. Our study reveals that the deviations are primarily caused by a small set of hard tokens scattered along the rationale sequence.

To give an intuitive illustration, we conduct an analytic experiment on MATH~\citep{hendrycks_measuring_2021} dataset that measures the disparity between a small LLM $\phi$ (Qwen2.5-3B-Instruct) and a high-performance oracle $\theta$ (Qwen2.5-32B-Instruct). 
Specifically, we feed model $\phi$ with an output from model $\theta$ to make predictions at each position, collect the inconsistent ones, and continue generation from the inconsistent positions using $\phi$ to see whether it leads to an incorrect final answer (deviation).

\begin{figure}[t]
    \centering
    \includegraphics[width=0.9\linewidth]{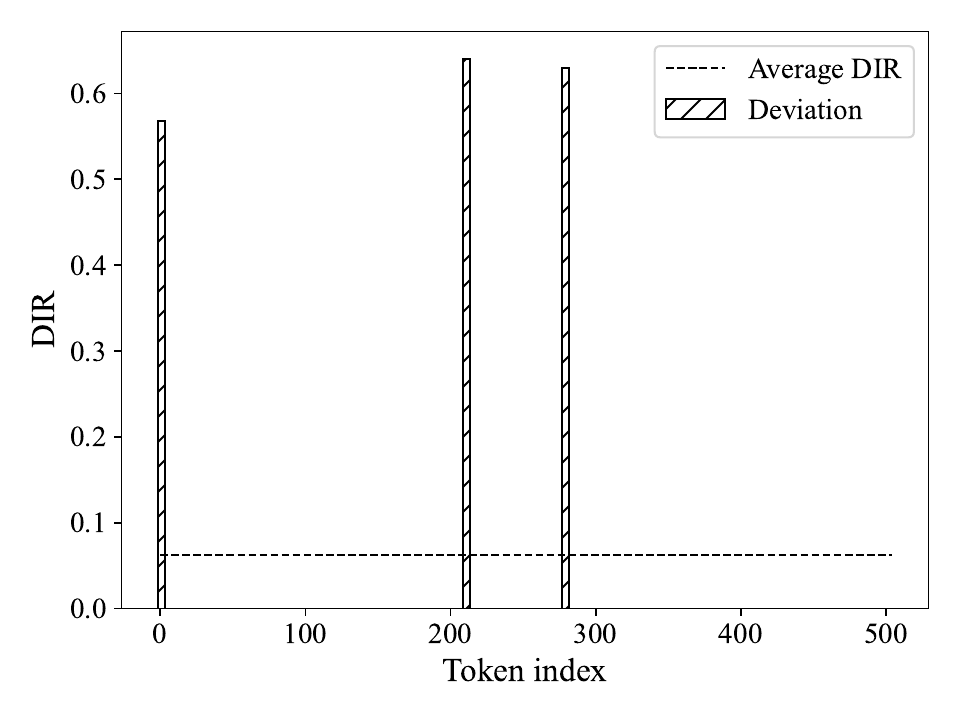}
    \caption{We depict the positions where Qwen2.5-3B-Instruct deviates from the correct rationale using the bars. Additionally, we annotate their DIR value by the heights, which significantly exceeds the average line, showing the indicative power of DIR.}
    \label{fig:pulse}
\end{figure}

Figure~\ref{fig:pulse} shows a representative case: among a generation of 500 tokens, only 3 of them cause model $\phi$ to deviate from the correct reasoning paths. These positions show high DIR values, inspiring the design of an intervention strategy.

\section{Hint-Practice Reasoning}
\begin{figure*}
    \centering
    \includegraphics[width=0.97\linewidth]{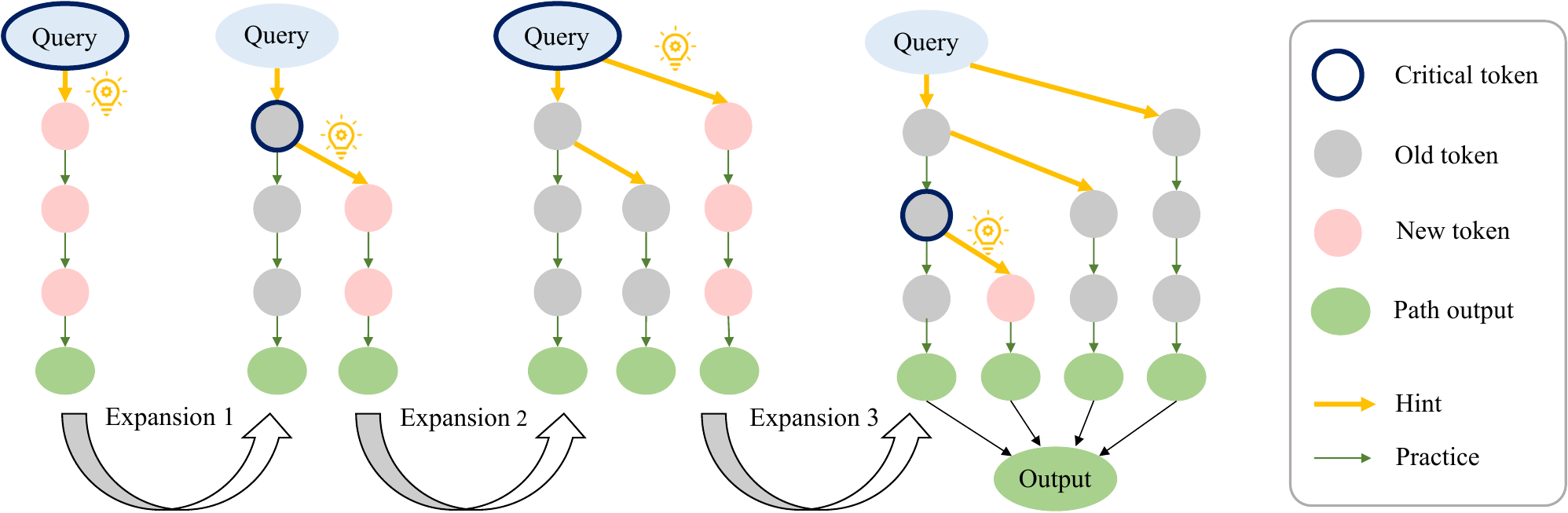}
    \caption{Illustration of the HPR iterative procedure. Each circular node denotes a token. HPR will generate multiple reasoning paths, starting by generating a single chain of thought. In each subsequent iteration, HPR selects a critical token (red circled) and expand a new path. This process repeats until the maximum iteration limit is reached.}
    \label{fig:overview}
\end{figure*}
We first introduce our targeted tree-structure decoding workflow in Section~\ref{sec:workflow}.
In Section 4.2, we introduce the core innovation of HPR---Distributional Inconsistency (DI)---a quantified measure where higher values indicate greater potential discrepancy between the current reasoning tree outcomes and the hinter model's expected results.
To minimize DI under constrained intervention budgets, Section 4.3 presents our methodology for computing Distributional Inconsistency Reduction (DIR), which estimates each node's potential contribution to DI minimization.

\subsection{Targeted Reasoning Tree Expansion}\label{sec:workflow}
HPR aims to allocate test-time computation to the reasoning state where it is truly needed.
We adopt a tree-structured search to progressively explore the thought space and reuse high-quality rationale prefixes, avoiding redundant computation. To aggregate reasoning paths, we apply weighted voting, a proven method for test-time scaling~\citep{zhang_what_2025, wang_self-consistency_2022}.

However, prior tree-based methods~\citep{yao_tree_2023, mo_tree_2024, hao_reasoning_2023} often generate many incomplete paths, which are excluded from final voting, leading to inefficiency. To address this, we ensure every expansion yields a complete rationale by letting the practitioner finish each branch.

HPR employs a ``grow-from-the-middle'' strategy guided by the hinter. As illustrated in Figure~\ref{fig:overview}, the tree is expanded iteratively through four phases:
\begin{itemize}
    \item \textbf{Select}: Compute DIR to identify the node where the hinter can contribute the most novel information; this becomes the \textit{critical node}.
    \item \textbf{Hint}: Feed the hinter with the critical node’s prefix to generate a hint.
    \item \textbf{Practice}: The practitioner completes the reasoning via greedy decoding based on the hint.
    \item \textbf{Analyze}: Record output distributions to support DIR calculation for the next iteration.
\end{itemize}

HPR iteratively augments the reasoning tree through strategic interventions to progressively align the tree with reasoning trees sampled from the hinter model's distribution.
To operationalize this objective, we need a metric for evaluating the global divergence between the current reasoning tree and the target distribution, which we will introduce in the next section. 

\subsection{Distributional Inconsistency}\label{dcg}
To enable the comparison between the reasoning tree and a distribution, we first need to establish a characterization distribution that describes the already explored trajectories. Furthermore, since we have access to a more accurate distribution, we can incorporate hinter's information into the characterization distribution to make preliminary rectification.
\begin{definition}[Characterization Distribution]
    Given a query $\bm x$, a hinter model $\theta$, and a practitioner model $\phi$, suppose there have been $m$ reasoning paths sampled from $P_\phi$ of length $n$: $V=(\bm r_{1:n}^1, \bm r_{1:n}^2,\cdots,\bm r_{1:n}^m)$, allowing overlap at prefix, characterization distribution $Q_V$ is defined as:
    \begin{equation}
        Q_V(r_i|\bm x, \bm r_{1:i-1})=\frac{P_\theta(r_i|\bm x, \bm r_{1:i-1})}{\sum_{r'_i\in N_V(\bm r_{1:i-1})}P_\theta(\bm r'_i|r_{1:i-1})},
    \end{equation}
    where $N_V(\bm r_{1:i-1})$ denotes all successors of $\bm r_{1:i-1}$ in $V$.
\end{definition}

Intuitively, $Q_V$ is the distribution that embeds $P_\theta$ onto the support set $V$. It is built upon $P_\phi$ and reallocates the token probability mass towards $P_\theta$. Sampling from $Q_V$ is equivalent to sampling from $P_\theta$ while rejecting rationales not in $V$. HPR samples from $Q_V$ to get refined outputs by marginalizing over all explored reasoning paths:
\begin{equation}
    P^{HPR}(\bm y|\bm x;V)=\sum_{\bm r\in V}Q_V(\bm r|\bm x)P_\phi(\bm y|\bm x, \bm r).
\end{equation}

There are four features of $Q_V$: 
(1) The two components of $Q_V$---support set $V$ and probability value $P_\theta$---are disentangled, which facilitates the analysis of the collaboration between the two models; 
(2) $Q_V$ provides intermediate probabilistic information which helps guide the expansion of the reasoning tree; 
(3) $Q_V$ equals 1 at each non-branching node, simplifying computational implementation; 
(4) $Q_V$ enables the measurement of the divergence between the observed paths and hinter's distribution.
\begin{definition}[Distributional Inconsistency]
    The \textit{distributional inconsistency} between rationales $V$ and $P_\theta$ is defined as the reversed Kullback–Leibler divergence between surrogate sampling distribution and the hinter distribution: $D_{KL}(Q_V||P_\theta)$.
\end{definition}

$D_{KL}(Q_V||P_\theta)$ reflects how much valuable space has not been covered by the current reasoning tree. It provides a global search guidance as it considers the contribution of all tree nodes.

\subsection{Distributional Inconsistency Reduction}
We expect the $Q_V$ to be as close as possible to $P_\theta$. To achieve this, we adopt a greedy strategy that lets practitioner generate a rationale $\bm v$ maximizing the reduction of KL divergence $D_{KL}(Q_V||P_\theta)-D_{KL}(Q_{V\cup\{\bm v\}}||P_\theta)$ at each iteration.

\begin{figure}[t]
    \centering
    \includegraphics[width=0.7\linewidth]{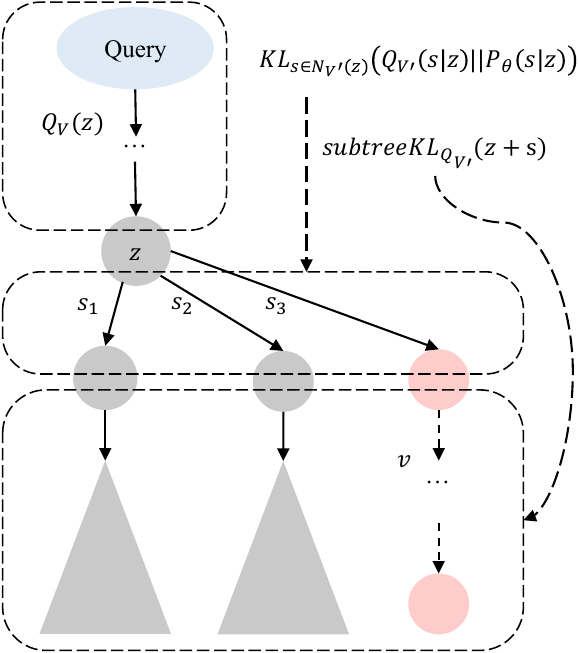}
    \caption{Illustration of three terms in node-version DIR. The pink circles denote the newly generated path.}
    \label{fig:metric}
\end{figure}

During the ``select'' phase, we need to select a node $\bm z$ such that the newly expanded branch from $\bm z$ yields a high reduction.
However, trajectory-based reduction cannot be efficiently evaluated for each node.
Therefore, we reorganized the calculation and derived the final node-based formulation of Distributional Inconsistency Reduction (DIR).
For conciseness, we omit $\bm x$ in the condition.
\begin{definition}[Distributional Inconsistency Reduction]
    Let $\bm z$ be a prefix of some reasoning path in $V$, $\bm v$ be the newly generated path starting from $\bm z$, $V'=V\cup\{\bm v\}$.
    \begin{equation}
        \begin{split}
            &{\rm DIR}(\bm z;V,P_\theta)=D_{KL}(Q_V||P_\theta)-D_{KL}(Q_{V\cup\{\bm v\}}||P_\theta)\\
            =\ &Q_V(\bm z)\biggl[\Delta D_{{KL}_{s\in N(\bm z)}}\left(Q(s|\bm z)||P_\theta(s|\bm z)\right)+\\
            &\sum_{s\in N_{V'}(\bm z)}\Delta Q(s|\bm z){\rm subtreeKL}_{Q_{V'}}(\bm z+s)\biggr],
        \end{split}
    \end{equation}
    where $\Delta D_{{KL}}$ denotes the difference between pre-expansion KL  $D_{{KL}_{s\in N_{V}}(\bm z)}\left(Q_V(s|\bm z)||P(s|\bm z)\right)$ and post-expansion KL $D_{{KL}_{s\in N_{V'}}(\bm z)}\left(Q_{V'}(s|\bm z)||P(s|\bm z)\right)$, $\Delta Q(s|\bm z)=Q_V(s|\bm z)-Q_{V'}(s|\bm z)$, $\bm z'=\bm z+s$ means concatenating prefix with token $s$,
    ${\rm subtreeKL}_{Q_{V'}}(\bm z')=\sum_{\bm z'\subset \bm v\in V}Q_{V'}(\bm v-\bm z'|\bm z') \log\frac{Q(\bm v-\bm z'|\bm z')}{P_\theta(\bm v-\bm z'|\bm z')}$.
\end{definition}

\begin{algorithm}[t]
   \caption{Hint-Practice Reasoning}
   \label{alg:hpr}
\begin{algorithmic}
   \REQUIRE practitioner $\phi$, hinter $\theta$
   \STATE {\bfseries Input:} query $\bm x$, max\_iterations $k$
   \STATE $P=V=A=\emptyset, nodes=\{\bm x\}$
   \REPEAT
   \STATE critical node $\bm z=\arg\max_{\bm z\in nodes}{\rm DIR}(\bm z;V,P)$
   \STATE hint $\bm h~\sim P_\theta(\cdot|\bm z)$
   \STATE practice $\bm p\sim P_\phi(\cdot|\bm {z+h})$
   \STATE $\bm v=\bm {h+p}$
   \STATE answer $a\sim P_\phi(\cdot|\bm {z+v})$
   \STATE $V=V\cup\{\bm z+\bm v\}$
   \STATE $A=A\cup\{a\}$
   \STATE $u_i=H(P_\phi(\cdot|\bm z+\bm v_{0:i}))$, for $i=1,2,\dots,len(\bm v)$
   \STATE $\bm v'=\bm v_{1:i}$, $i=\min\{i|Top\text{-}U(\{u_j\}_{j=1}^{len(\bm v)})\subset \bm u_{1:i}\}$
   \STATE $P'=\{P_i\}_{i=1}^{len(\bm v')}=Top\text{-}K(P_\theta(\bm v'|\bm z))$
   \FOR{$i=1$ {\bfseries to} $len(\bm v')$}
        \STATE $nodes=nodes\cup\{\bm z+\bm v_{1:i}\}$
        \STATE $P=P\cup P'$
   \ENDFOR
   \UNTIL{reach max iterations}
   \RETURN Weighted-vote($A,Q_V$)
\end{algorithmic}
\end{algorithm}

Details of the derivation are included in Appendix.
The three main terms in DIR correspond to contributions from the prefix, next-token, and sub-tree, respectively, as illustrated in Figure~\ref{fig:metric}. These terms reflect three characteristics of a good expansion starting point $z$:
\begin{itemize}
    \item DIR favors prefixes with high probability, ensuring that new paths are built on a solid and reasonable foundation.
    \item DIR favors those under-explored nodes. The next-token term is particularly significant when there is a large hinter-practitioner gap or when the practitioner exhibits uncertainty ($\Delta D_{{KL}_{s\in N(\bm z)}}\left(Q(s|\bm z)||P_\theta(s|\bm z)\right)$), aligning with our core motivation.
    \item DIR favors the expansion where the path $\bm v-\bm z'$ is likely to occur (the subtreeKL of a chain equals to the sum of $-\log P_\theta(\cdot)$).
\end{itemize}

In our implementation, we normalize the subtree KL by dividing it by the average path length among the sub-tree to avoid DIR being biased. During the ``analyze'' phase, HPR will evaluate and store probabilities of Top-K next tokens ($K=32$ in experiments) for each expanded node. The only unknown term in DIR computation is the probability of new path $\bm v-\bm z'$, as it must be calculated before expansion. As an approximation, We use the average log probabilities of the nearest 32 tokens in the prefix and the sibling greedy decoding path.

To calculate DIR, we need to leverage the hinter to re-evaluate the rationales to collect $P_\theta$. This process requires only a single forward operation to obtain all necessary probabilities, avoiding token-by-token decoding, thus introducing minimal latency.
To be more efficient, for newly generated paths, we restrict the selection of critical tokens to the minimal prefix that contains the Top-$U$ most uncertain tokens with regard to $P_\phi$ (HPR uses information entropy $H$) because we empirically find a strong correlation between $P_\phi$ uncertainty and DIR value.
In practice, we find that the performance and the efficiency are insensitive to $U$. We set $U=3$ in all experiments. The complete procedure of HPR is presented in Algorithm~\ref{alg:hpr}.

\section{Experiments}

\setlength\tabcolsep{3pt}
\begin{table*}[t]
    \centering
    \footnotesize
    \begin{tabular}{lccccccccccc}
        \toprule
        \multirow{2}{*}{\textbf{Method}}&\multicolumn{3}{c}{\textbf{Arithmetic}}&&\multicolumn{2}{c}{\textbf{Commonsense}}&&\multicolumn{3}{c}{\textbf{Inference Cost}}&\multirow{2}{*}{\textbf{\shortstack{Efficiency\\(REE)}}}\\
        \cmidrule{2-4}\cmidrule{6-7}\cmidrule{9-11}
        &GSM8K&AQUA-RAT&MATH&&\makebox[4em][c]{CSQA}&\makebox[4em][c]{StrategyQA}&&\#Tokens (P)&\#Tokens (H)&FLOPs ($10^{12}$)&\\
        \midrule
        CoT&85.3&64.2&53.0&&74.5&59.5&&\dbox{320.8}&-&\dbox{1.6}&-\\
        CoT-SC@5&88.9&69.2&59.8&&76.1&59.9&&\dbox{1664.8}&-&\dbox{8.4}&\dbox{0.82}\\
        CoT-SC@15&90.2&73.2&\textbf{63.4}&&78.4&60.2&&\dbox{4720.8}&-&\dbox{23.6}&\dbox{0.42}\\
        ToT@5&84.8&68.2&53.4&&75.0&61.7&&\dbox{6916.8}&-&\dbox{34.6}&\dbox{0.06}\\
        TouT@5&85.3&71.2&53.6&&74.9&62.0&&\dbox{6873.0}&-&\dbox{34.4}&\dbox{0.10}\\
        MCTS@5&87.4&69.7&58.9&&75.1&60.0&&\dbox{5714.6}&-&\dbox{28.6}&\dbox{0.17}\\
        \midrule
        \multicolumn{12}{c}{\textit{Guided by Qwen2.5-14B-Instruct}} \\
        \midrule
        Hinter-SC@5 (UB)&94.2&84.5&72.7&&84.0&75.8&&-&\dbox{1676.4}&\dbox{45.2}&\dbox{0.55}\\
        CoT-BoN@5&88.1&69.3&56.3&&76.1&61.6&&\dbox{1664.8}&\dbox{1664.8}&\dbox{19.6}&\dbox{0.26}\\
        AdaSwitch&89.9&69.3&59.7&&75.0&60.5&&\dbox{1013.8}&\dbox{185.0}&\dbox{10.0}&\dbox{0.68}\\
        HPR@5 (Ours)&\textbf{91.0}&\textbf{73.2}&62.1&&78.0&\textbf{62.0}&&\dbox{936.8}&\dbox{124.2}&\dbox{8.0}&\dbox{\textbf{1.49}}\\
        \midrule
        \multicolumn{12}{c}{\textit{Guided by Qwen2.5-32B-Instruct}} \\
        \midrule
        Hinter-SC@5 (UB)&95.2&87.0&75.4&&86.0&76.1&&-&\dbox{1638.0}&\dbox{103.2}&\dbox{0.26}\\
        CoT-BoN@5&88.6&69.7&56.8&&76.2&62.0&&\dbox{1664.8}&\dbox{1664.8}&\dbox{34.6}&\dbox{0.16}\\
        AdaSwitch&89.9&68.5&60.6&&75.0&60.7&&\dbox{1014.4}&\dbox{171.4}&\dbox{15.8}&\dbox{0.41}\\
        HPR@5 (Ours)&\textbf{91.8}&\textbf{74.8}&63.2&&\textbf{78.9}&\textbf{63.6}&&\dbox{975.5}&\dbox{124.1}&\dbox{12.8}&\dbox{\textbf{1.02}}\\
        \bottomrule
    \end{tabular}
    \caption{Results using Qwen2.5-3B-Instruct as practitioner. ``@$k$'' means that $k$ reasoning paths are sampled during inference. \#Tokens (P)/(H) denote the number of tokens generated from practitioner and hinter, respectively (For BoN, \#Tokens (H) denotes the number of evaluated tokens). UB means the accuracy upper bound within the same group. We mark a result bold if it outperforms all practitioner-only baselines and guided-search baselines within the same group. Cost and efficiency are averaged across 5 datasets.}\label{tab:3b}
\end{table*}

\begin{table*}[t]
    \centering
    \footnotesize
    \begin{tabular}{lccccccccccc}
        \toprule
        \multirow{2}{*}{\textbf{Method}}&\multicolumn{3}{c}{\textbf{Arithmetic}}&&\multicolumn{2}{c}{\textbf{Commonsense}}&&\multicolumn{3}{c}{\textbf{Inference Cost}}&\multirow{2}{*}{\textbf{\shortstack{Efficiency\\(REE)}}}\\
        \cmidrule{2-4}\cmidrule{6-7}\cmidrule{9-11}
        &GSM8K&AQUA-RAT&MATH&&\makebox[4em][c]{CSQA}&\makebox[4em][c]{StrategyQA}&&\#Tokens (P)&\#Tokens (H)&FLOPs ($10^{12}$)&\\
        \midrule
        CoT&90.8&80.0&66.8&&81.5&69.6&&\dbox{327.4}&-&\dbox{4.0}&-\\
        CoT-SC@5&93.1&83.9&71.8&&82.1&71.5&&\dbox{1672.6}&-&\dbox{21.8}&\dbox{0.62}\\
        CoT-SC@15&93.4&\textbf{87.0}&\textbf{75.3}&&82.6&72.0&&\dbox{5042.2}&-&\dbox{65.6}&\dbox{0.28}\\
        ToT@5&89.7&83.7&66.1&&80.5&70.1&&\dbox{7542.4}&-&\dbox{98.0}&\dbox{0.01}\\
        TouT@5&90.1&82.2&66.5&&81.1&71.0&&\dbox{7548.0}&-&\dbox{98.0}&\dbox{0.02}\\
        MCTS@5&91.6&84.6&71.7&&81.1&71.4&&\dbox{6142.6}&-&\dbox{79.8}&\dbox{0.12}\\
        \midrule
        \multicolumn{12}{c}{\textit{Guided by Qwen2.5-14B-Instruct}} \\
        \midrule
        Hinter-SC@5 (UB)&94.2&84.5&72.7&&84.0&75.8&&-&\dbox{1676.4}&\dbox{43.2}&\dbox{0.43}\\
        CoT-BoN@5&92.0&84.6&70.4&&81.1&71.9&&\dbox{1672.6}&\dbox{1672.6}&\dbox{33.0}&\dbox{0.31}\\
        AdaSwitch&92.6&85.0&69.2&&82.1&71.5&&\dbox{1055.2}&\dbox{140.8}&\dbox{17.6}&\dbox{0.69}\\
        HPR@5 (Ours)&93.0&85.5&71.8&&\textbf{82.7}&\textbf{72.5}&&\dbox{936.8}&\dbox{124.0}&\dbox{15.6}&\dbox{\textbf{1.16}}\\
        \midrule
        \multicolumn{12}{c}{\textit{Guided by Qwen2.5-32B-Instruct}} \\
        \midrule
        Hinter-SC@5 (UB)&95.2&87.0&75.4&&86.0&76.1&&-&\dbox{1638.0}&\dbox{103.2}&\dbox{0.25}\\
        CoT-BoN@5&92.5&84.6&70.6&&81.4&72.4&&\dbox{1672.6}&\dbox{1672.6}&\dbox{48.0}&\dbox{0.23}\\
        AdaSwitch&93.0&85.8&69.7&&82.7&71.5&&\dbox{1045.6}&\dbox{127.6}&\dbox{21.6}&\dbox{0.64}\\
        HPR@5 (Ours)&\textbf{93.6}&86.2&72.7&&\textbf{83.4}&\textbf{73.4}&&\dbox{944.4}&\dbox{124.2}&\dbox{20.0}&\dbox{\textbf{1.03}}\\
        \bottomrule
    \end{tabular}
    \caption{Results using Qwen2.5-7B-Instruct as practitioner.}\label{tab:7b}
\end{table*}

\subsection{Setup}

\paragraph{Benchmarks}
We evaluate our approach on arithmetic reasoning and commonsense reasoning tasks. Arithmetic reasoning includes GSM8K~\citep{cobbe_training_2021}, AQUA-RAT~\citep{chen_program_2023}, and MATH~\citep{hendrycks_measuring_2021}. These datasets are among the more challenging ones used in prior studies~\citep{sun_adaswitch_2024, ma_non-myopic_2025}. For commonsense reasoning, we select CSQA~\citep{talmor_commonsenseqa_2019} and StrategyQA~\citep{geva_did_2021}, both of which require implicit reasoning.

\paragraph{Implementation}
We use models from Qwen2.5 series~\citep{qwen2.5}: 3B/7B for practitioner and 14B/32B for hinter. To account for differences in task complexity, we set the HPR hint length to 32 tokens for arithmetic reasoning and 16 tokens for commonsense reasoning. We use vLLM~\citep{kwon_efficient_2023} library to provide efficient inference serving. For all experiments, we use ``Let's think step by step'' prompting~\citep{kojima_large_2022} to perform zero-shot CoT. We set the temperature to 0.7 for all non-greedy generations and average results across three different seeds.

\paragraph{Baselines}
We compare HPR with single-trajectory CoT, sampling-based method Self-Consistency (SC)~\citep{wang_self-consistency_2022}, tree-structure search methods ToT~\citep{yao_tree_2023}, TouT~\citep{mo_tree_2024}, and MCTS~\citep{hao_reasoning_2023}, as well as externally guided baselines Best-of-N (BoN)~\citep{song_good_2024} and AdaSwitch~\citep{sun_adaswitch_2024}.
Besides, we report a pure hinter baseline to show the performance upper bound of externally guided methods.
We use the hinter as the reward model both for BoN and MCTS. For MCTS, we follow Marco-o1~\citep{zhao_marco-o1_2024} to use the average confidence as the reward score, and ensemble all paths via majority voting. ToT, TouT, and MCTS are allowed to sample 5 different actions at each step in our experiments. The hyperparameters for AdaSwitch are tuned to ensure a comparable token budget with HPR.

\paragraph{Metric}
We extract the final answer from the LLM's output and calculate exact match accuracy.
To distinguish the contributions of the practitioner and the hinter, we recorded the average number of tokens involved by each during inference (averaged across samples).
We calculate the FLOPs (floating point operations) to measure the cost of different approaches.
Following \citet{kaplan_scaling_2020}, the FLOPs of decoding a sequence of length $L$ approximately equal $2NL$, where $N$ is the number of non-embedding parameters of the LLM.
To evaluate the reasoning-expansion efficiency (REE), we devise an integrated measure by computing the normalized gain-cost ratio:
\begin{equation}
    REE=\frac{ACC-A_0}{FLOPs-F_0} * \frac{F_0}{100\%},
\end{equation}
where $A_0$ and $F_0$ are the accuracy and FLOPs of single CoT baseline.
Given that decoding operation is typically more time-consuming than evaluation due to bandwidth limitations, we follow API cost conventions and weight evaluation FLOPs by a factor of 1/4 for fair comparison.

\subsection{Main Results}
The results of Qwen2.5-3B/7B-Instruct are shown in Table~\ref{tab:3b} and Table~\ref{tab:7b}, which represent the settings of high/low practitioner-hinter capability gap, respectively. We can make several observations:

First, HPR shows the highest token and computational efficiency. HPR@5 consumes only about 2/3 tokens of CoT-SC@5 and maintains similar FLOPs, though it involves a larger model. CoT-SC strategy has to generate 5x tokens to reach similar performance. When compared with other tree search methods, the advantage of token efficiency and FLOPs increases to 3x-5x, which demonstrates that our targeted tree expansion managed to avoid wasteful generation.
Notably, hinter is involved in very little generation, which means deploying one hinter can simultaneously provide service for multiple practitioners, saving high-memory devices.

Second, HPR shows consistent improvement over all baselines across all benchmarks. It outperforms sampling and tree-structured baselines by 5.1\% at most when searching the same number of reasoning paths while costing less computation, which demonstrates that introducing the aid of a stronger model is an economical way to boost performance. Compared with BoN and AdaSwitch, HPR also achieves an improvement of 5.1\% at most, which we attribute to the effective exploration mechanism in HPR.

Third, HPR's performance is close to the pure hinter upper bound in the low-capability-gap setting while consuming 1/5-1/3 FLOPs. This indicates that, in certain cases, HPR could be a more efficient alternative compared to deploying a single strong model.

\begin{figure}[t]
    \centering
    \includegraphics[width=0.9\linewidth]{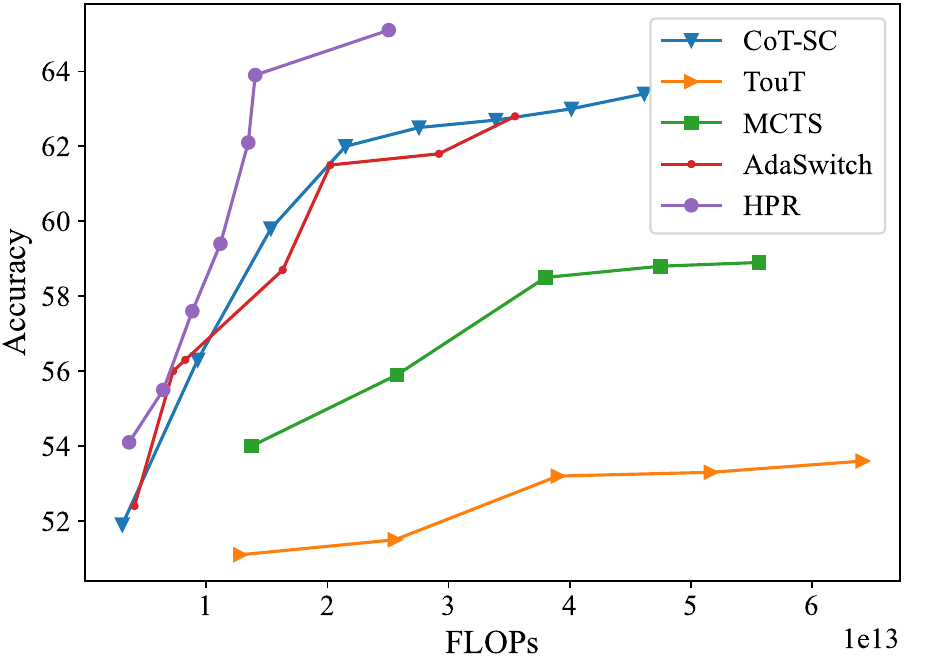}
    \caption{Accuracy versus FLOPs per instance on the MATH dataset using Qwen2.5-3B/14B-Instruct with varying numbers of reasoning paths.}
    \label{fig:effi}
\end{figure}

\subsection{Performance-Cost Analysis}
Previous experiments primarily focus on the 5-trajectory setting. In this section, we comprehensively compare HPR with baseline methods under different computational budgets by varying the number of trajectories. The detailed experimental settings are provided in Table 1 in the Appendix. As shown in Figure~\ref{fig:effi}, HPR consistently achieves the highest accuracy across all FLOPs levels, demonstrating the general effectiveness of our DIR-guided tree expansion strategy.

\begin{figure}[t]
    \centering
    \includegraphics[width=0.9\linewidth]{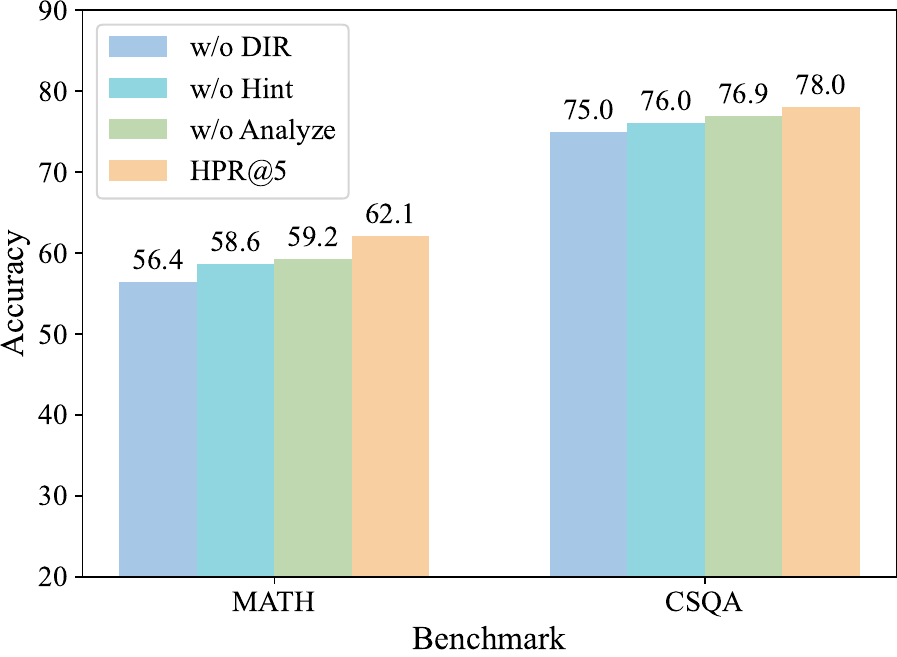}
    \caption{Results of ablation study using Qwen2.5-3B/14B-Instruct.}
    \label{fig:ablation}
\end{figure}

\begin{table*}[t]
    \centering
    \begin{tabular}{p{0.93\linewidth}}
        \toprule
         \textbf{Problem:} Find the maximum volume of a cone that fits inside a sphere of radius 1.\\
         \textbf{Initial trajectory:} Let's think step by step. We want to find the maximum volume of a cone that fits inside a sphere of radius 1. The volume \( V \) of a cone is given by the formula: $V = \frac{1}{3} \pi r^2 h$ where \( r \) is the radius of the base of the cone and \( h \) is the height of the cone. For a cone inscribed in a sphere, the relationship between the radius of the sphere, the radius of the base of the cone, and the height of the cone can be derived using the Pythagorean theorem. We have: $1^2 = r^2 + (h - \textcolor{red}{r)^2.}$ \textcolor{red}{Expanding and simplifying the equation: ...}\\
         \textbf{Expansion 1:} \textcolor{gray}{Let's think step by step ... So, we have: $1^2 = r^2 + (h -$} \textcolor{blue}{$1)^2$. This simplifies to:$r^2 = 2h - h^2$.} Now, substitute \( r^2 = 2h - h^2 \) into the volume formula: $V = \frac{1}{3} \pi (2h - h^2) h = \frac{1}{3} \pi (2h^2 - h^3)$. To find the maximum volume, we need to take the derivative of \( V \) with respect to \( h \) and set it to zero: $\frac{dV}{dh} = \frac{1}{3}$, which gives $\pi (4h - 3h^2)$, $h(4 - 3h) = 0$. This gives us two solutions: \( h = 0 \) (which is not physically meaningful for a cone) and \( h = \frac{4}{3} \). \textcolor{red}{However, since \( h = \frac{4}{3} \) is outside the valid range for a cone inscribed in a sphere of radius 1, ...}\\
         \textbf{Expansion 2:} \textcolor{gray}{Let's think step by step ... and \( h = \frac{4}{3} \)}. \textcolor{blue}{We need to check if this value of \( h \) gives a maximum volume.} We can do this by checking the second derivative ... So, the answer is $\frac{32\pi}{81}$.\\
         \bottomrule
    \end{tabular}
    \caption{Case study of solving math problem. Red text indicates the mistake of practitioner. Blue text indicates the hint. Gray text indicates reused prefix.}\label{tab:case}
\end{table*}

\subsection{Ablation Study}
HPR experiences four phases during one iteration. In this section, we will examine the effect of these phases by three variants: (1) disable DIR and randomly select from all non-terminal nodes with equal probabilities; (2) remove the hint generation and direct call practitioner to continue generation from selected critical node; (3) skip the ``analyze'' phase and use practitioner's probability to calculate $Q_V$.

The results are shown in Figure~\ref{fig:ablation}. We can find that eliminating any phase of HPR leads to a significant performance drop.
Among them, the most important one is DIR-guided selection, as it ensures the hinter provides guidance at the most critical points, forming the foundation of the HPR framework. For the variant without hint, the practitioner knows where to expand from but struggles to decode the correct token. The ``w/o Analyze'' performs slightly better that other two, as it retains most of HPR’s mechanisms but shifts alignment toward the practitioner’s inherent probabilities.

\begin{figure}
    \centering
    \includegraphics[width=0.9\linewidth]{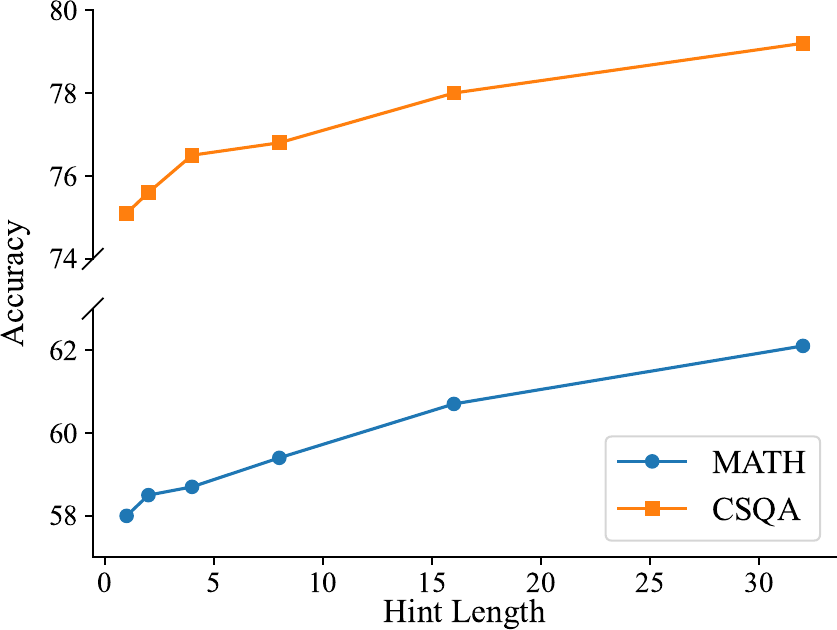}
    \caption{Performance of HPR@5 with varying hint length using Qwen2.5-3B/14B-Instruct.}
    \label{fig:hint}
\end{figure}

\subsection{Impact of Hint Length}
The hint length determines the extent to which the hinter contributes to the generation process. To analyze its impact, we test different hint lengths ranging from 1 to 32. The results are shown in Figure~\ref{fig:hint}.

We observe the most significant performance gain when increasing the hint length from 1 to 4 tokens. Beyond this point, the improvement continues at a slower, approximately linear rate. These results suggest that hinter's primary role is to guide the reasoning direction—demonstrating that even short hints can be highly effective. Moreover, longer hints consistently enhance performance, gradually pushing the practitioner’s output closer to that of the hinter model.

\subsection{Case Study}
We examine examples of math problem solving to illustrate the effectiveness of HPR. As shown in Table~\ref{tab:case}, the initial rationale fails to derive the correct equation.
The hinter identifies and corrects the wrong token ``$r$'', giving practitioner a second chance.
Subsequently, the practitioner makes another mistake regarding the validity of $h$, and the hinter again intervenes to adjust the reasoning direction.
Finally, the practitioner successfully arrives at the correct answer.
This case demonstrates that timely and strategic interventions from the hinter can effectively and efficiently steer the reasoning process toward the correct solution.

\section{Conclusion}
In this paper, we propose a novel reasoning framework (HPR) that enables timely and efficient intervention during decoding to overcome challenging ``critical tokens'' and correct the reasoning direction.
HPR is theoretically backed up by our newly proposed Distributional Inconsistency Reduction that measures the disparity between the already explored trajectories and the targeted hinter distribution.
HPR leverages DIR to guide the reasoning expansion node selection and iteratively expand the reasoning tree.
Extensive experiments on arithmetic and commonsense reasoning tasks demonstrate the effectiveness of HPR, highlighting its ability to enhance reasoning accuracy while maintaining computational efficiency.

\section{Future Work}
Since HPR provides an elegant framework to integrate the advantages of two models, it inspires us to generalize the model combination beyond the reasoning dimension.
For example, a promising direction to use a general-purpose hinter to help a domain-specific practitioner better tackle complex tasks.
In addition, we are studying the vocabulary alignment for DIR computation to support the model combinations from different model families.

\bibliography{aaai2026}

\appendix

\begin{table}[b]
    \centering
    \begin{tabular}{l|l}
        \toprule
        Method & Number of Paths \\
        \hline
        CoT-SC & 1,3,4,5,7,9,11,13,15\\
        TouT & 1,2,3,4,5 \\
        MCTS & 1,2,3,4,5 \\
        AdaSwitch & 1,2,3,4,5 \\
        HPR & 1,2,3,4,5,7 9\\
        \bottomrule
    \end{tabular}
    \caption{Number of reasoning paths for each method in the efficiency analysis.}
    \label{tab:effi_setting}
\end{table}

\section{Derivation of Node DIR}
\label{apsec:dcg}
The main idea of the derivation is to decompose the KL-divergence following the tree structureaccording to the tree structure, allowing us to cancel out identical terms. When a new path is expanded from node $\bm z$, the only updated terms of $Q_{V'}$ are $Q_{V'}(\cdot|\bm z)$. So, we only need to consider paths that contain $\bm z$. Additionally, the update of $Q_{V'}(\cdot|\bm z)$ does not influence the value of the prefix value $Q_V(\bm z)$ or the nodes in the sub-tree of $\bm z$'s children. This allows we to further split each KL term into three parts: prefix, child, and subtree. By grouping the each part, we drive the final formulation of Node DIR. The derivation is provided in the next page.

\begin{figure*}[t]
    \centering
    \begin{align*}
        &\text{DIR}(z;V,P)\\
        =\ &D_{KL}(Q_V||P)-D_{KL}(Q_{V'}||P)\\
        =\ &\sum_{\bm v\in \bm v s.t. \bm z\subset V}Q_V(\bm v)\log\frac{Q_V(\bm v)}{P(\bm v)} - \sum_{\bm v\in V' s.t. z\subset\bm v}Q_{V'}(\bm v)\log\frac{Q_{V'}(\bm v)}{P(\bm v)}\\
        =\ &Q_V(\bm z)\left[\sum_{\bm v\in V s.t. z\subset\bm v}Q_V(\bm v-\bm z|\bm z)\log\frac{Q_V(\bm v-\bm z|\bm z)Q_V(\bm z)}{P(\bm v-\bm z|\bm z)P(\bm z)}\right.\\
        \ &\left.-\sum_{\bm v\in V' s.t. z\subset\bm v}Q_{V'}(\bm v-\bm z|\bm z)\log\frac{Q_{V'}(\bm v-\bm z|\bm z)Q_{V'}(\bm z)}{P(\bm v-\bm z|\bm z)P(\bm z)}\right]\\
        =\ &Q_V(\bm z)\left[\sum_{\bm v\in V s.t. z\subset\bm v}Q_V(\bm v-\bm z|\bm z)\log\frac{Q_V(\bm v-\bm z|\bm z)}{P(\bm v-\bm z|\bm z)}-\sum_{\bm v\in V' s.t. z\subset\bm v}Q_{V'}(\bm v-\bm z|\bm z)\log\frac{Q_{V'}(\bm v-\bm z|\bm z)}{P(\bm v-\bm z|\bm z)}\right]\\
        =\ &Q_V(\bm z)\left[\sum_{\substack{s\in N_V(\bm z)\\\bm v\ s.t. s\in\bm v}}Q_V(\bm v-\bm z-s|\bm z+s)\log\frac{Q_V(\bm v-\bm z-s|\bm z+s)Q_V(s|\bm z)}{P(\bm v-\bm z-s|\bm z+s)P(s|\bm z)}\right.\\
        &\left.-\sum_{\substack{s\in N_V(\bm z)\\\bm v\ s.t. s\in\bm v}}Q_{V'}(\bm v-\bm z-s|\bm z+s)\log\frac{Q_{V'}(\bm v-\bm z-s|\bm z+s)Q_{V'}(s|\bm z)}{P(\bm v-\bm z-s|\bm z+s)P(s|\bm z)}\right]\\
        =\ &\text{\small$Q_V(\bm z)\left[\sum_{s\in N_V(\bm z)}Q_V(\bm s|\bm z)\left(\log\frac{Q_V(s|\bm z)}{P(s|\bm z)}+\sum_{\bm v\ s.t. s\in\bm v}Q_V(\bm v-\bm z-s|\bm z+s)\log\frac{Q_V(\bm v-\bm z-s|\bm z+s)Q_V(s|\bm z)}{P(\bm v-\bm z-s|\bm z+s)P(s|\bm z)}\right)\right.$}\\
        &\text{\small$\left.-\sum_{s\in N_{V'}(\bm z)}Q_{V'}(\bm s|\bm z)\left(\log\frac{Q_{V'}(s|\bm z)}{P(s|\bm z)}+\sum_{\bm v\ s.t. s\in\bm v}Q_{V'}(\bm v-\bm z-s|\bm z+s)\log\frac{Q_{V'}(\bm v-\bm z-s|\bm z+s)Q_{V'}(s|\bm z)}{P(\bm v-\bm z-s|\bm z+s)P(s|\bm z)}\right)\right]$}\\
        =\ &Q_V(\bm z)\left[\sum_{s\in N_V(\bm z)}Q_V(\bm s|\bm z)\left(\log\frac{Q_V(s|\bm z)}{P(s|\bm z)}+\text{subtreeKL}_{Q_V}(\bm z+s)\right)\right.\\
        &\left.-\sum_{s\in N_{V'}(\bm z)}Q_{V'}(\bm s|\bm z)\left(\log\frac{Q_{V'}(s|\bm z)}{P(s|\bm z)}+\text{subtreeKL}_{Q_{V'}}(\bm z+s)\right)\right]\\
        =\ &Q_V(\bm z)\left[\left(D_{KL}(Q_V(s|\bm z)||P(s|\bm z))+\sum_{s\in N_V(\bm z)}Q_V(s|\bm z)\text{subtreeKL}_{Q_V}(\bm z+s)\right)\right.\\
        &\left.\left(D_{KL}(Q_{V'}(s|\bm z)||P(s|\bm z))+\sum_{s\in N_V(\bm z)}Q_{V'}(s|\bm z)\text{subtreeKL}_{Q_{V'}}(z+\bm s)\right)\right]\\
        =\ & Q_V(\bm z)\left[\Delta D_{KL}(Q(s|\bm z)||P(s|\bm z))+\sum_{s\in N_V(\bm z)}\Delta Q(s|\bm z)\text{subtreeKL}_{Q_{V'}}(z+\bm s)\right]
    \end{align*}
\end{figure*}

\section{Setup of Efficiency Analysis}
\label{apsec:effi}
We control the FLOPs of each method by varying the number of reasoning paths from 1 to 15 (sampling size).
Because TouT, MCTS, AdaSwitch, and HPR require much more FLOPs compared with CoT-SC for sampling size, we restrict the maximum path number to 5 (to 9 for HPR) to avoid mismatched comparison.
Detailed setting are as shown in Table~\ref{tab:effi_setting}.

\end{document}